\documentclass[journal]{IEEEtran}

\usepackage[american]{babel}

\usepackage{natbib} 
\usepackage{mathtools} 
\usepackage{booktabs} 
\usepackage{tikz} 

\title{A Greedy Approach for  Offering to Telecom Subscribers}

\author{Piyush Kanti Bhunre, Tanmay Sen, and Arijit Sarkar

\IEEEcompsocitemizethanks {\IEEEcompsocthanksitem Piyush Kanti Bhunre, Tanmay Sen and Arijit Sarkar are with Ericsson, Kolkata, India
\{piyush.kanti.bhunre, tanmay.sen, arijit.sarkar\}@ericsson.com}}

\newtheorem{obs}{Observation}
\newtheorem{theorem}{Theorem}

\usepackage[utf8]{inputenc}
\usepackage{amsmath}
\usepackage{amssymb}

\begin{document}
\maketitle
\begin{abstract}
Customer retention or churn prevention is a challenging task of a telecom operator. One of the effective approaches is to offer some attractive incentive or additional services or money to the subscribers for keeping them engaged and make sure they stay in the operator's network for longer time. Often, operators allocate certain amount of monetary budget to carry out the offer campaign. 
The difficult part of this campaign is the selection of a set of customers from a large subscriber-base and deciding the amount that should be offered to an individual so that operator's objective is achieved. There may be multiple objectives (e.g., maximizing revenue, minimizing number of churns) for selection of subscriber and selection of an offer to the selected subscriber. Apart from monetary benefit, offers may include additional data, SMS, hots-spot tethering, and many more. This problem is known as offer optimization. In this paper, we propose a novel combinatorial algorithm for solving offer optimization under heterogeneous offers by maximizing expected revenue under the scenario of subscriber churn, which is, in general, seen in telecom domain. The proposed algorithm is efficient and accurate even for a very large subscriber-base.
\end{abstract}

\section{Introduction}\label{se:intro}
Offer optimization or campaign management is one of the routine tasks of a telephone operator for customer retention and service adoption.
A telecom operator always looks out for potential subscribers who may be interested for adopting certain  service in exchange of an incentive from the telecom operator. To achieve a business objective, an operator's task is to identify as set of subscribers and a set of appropriate offers that may be accepted by  the chosen  subscribers. Most of the time, this decision is made through a rule-based system which implement certain business rules and intuitive human judgment. \cite{resende2008handbook} describe varieties of optimization problems arise in telecommunication systems. \cite{johnson2013whom} formulate a constrained Optimization problem for maximizing the profit to a manufacturer by giving discounts to the customers.  
\cite{pham2021recommendation} proposed a offers recommender system for telecommunications.  \cite{cohen2004exploiting} addressed the problem of bank's marketing campaign optimization by solving a mixed integer programming (MIP) problem. \cite{nobibon2011optimization} proposed a branch-and-price algorithm to allocate one or more offers to the clients. \cite{verma2020offer} proposed a two stage framework for retail offer optimization, firstly, they have exploited generalized non-linear model based on temporal convolutional network to estimate item purchase probability, and secondly offer values are optimized by solving constrained based optimization problem with derived purchase probability. 

In this paper, we propose a novel greedy algorithm for solving this problem by maximizing expected revenue under the scenario of subscriber-churn from the network. In  this variant of the offer optimization problem, apart from a  monetary incentive given to the subscribers, the operator may award some other offers to the subscribers such as extra amount of data,  talk time, increased limit of  hots-spot data usage, unlimited data usage in certain time period of a day or week,  download, etc. 
The objective is to allocate appropriate offer to the subscribers who are most likely to accept the offer and  active in the network for a longer time period.  
The problem under consideration will be referred to as an {\em Offer Optimization Problem} (OOP). The contribution of our work is summarized below.
\begin{itemize}
\item We have proposed  a novel algorithm for finding optimum solution to the underlying optimization problem which our main contribution. 
The proposed algorithm is able to handles large number of subscribers. 
It provides optimal solution efficiently and out performs many existing algorithms which is evident from a comparison test given in the experiment section.  We also provided a theoretical argument why our algorithm provides accurate and  efficiency solution.
\item The algorithm is not limited to the offer optimization to telecom domain, but also applicable in solving problems coming from various other domains.

\end{itemize}

The rest of the paper is organized as follow. 
In Section~\ref{se:prob}, we present the mathematical problem formulation  which leads to solving the telecom OOP. 
The optimization algorithm along with necessary data structures, complexity analysis and relevant discussions are presented in Section~\ref{se:algo}. 
Section~\ref{se:exp} is furnished with experimental results and comparisons with  a couple of standard algorithms to validate the novelty and merit of the proposed algorithm.
Finally we conclude in Section~\ref{se:con}.


\section{Problem Formulation}\label{se:prob}
First we shall formulate the OOP under the incentives with different denominations. 
The same formulation is enough for solving more general problem with heterogeneous offers, where a offer may be monetary or  non-monetary services or advantages to the subscribers.
This generalization can be done by a simple transformation or  interpretation of the non-monetary offers which we shall discuss later part of this section.
Here, we assume that any of the existing subscribers may churn out from the network with some probability.
Further, each subscriber or a group of subscribers has some susceptibility towards an offer, which will be referred to as offer acceptance  rate.
The offer acceptance rate is used to model the probability of accepting an offer by a subscriber, which is incorporated in the objective function.

Suppose $S$ is a  set of $n$ subscribers, and the $i^\text{th}$ subscriber possesses the following characteristics:
\begin{itemize}
\item $p_i$: monthly top-up done by the subscriber
\item $\alpha_i$: probability that the subscriber may churn out
\item $\gamma_i$: acceptance rate, a parameter that defines sensitivity of $i^\text{th}$ subscriber towards an offer
\item $\beta_i$: probability of accepting an offer, and it is dependent on  $\gamma_i$.
\end{itemize}
The parameters  $\alpha_i$'s are estimated by using ML-based models and the subscriber's profile and usage data.
The rate of acceptance $\beta_i$'s are estimated from  data related to past offer campaigns and acceptance as well as subscriber profile.
The monthly top-up amount $p_i$'s comes from the rate plan and top-up history of the subscribers.

Note that for a large segment of subscribers, if the amount offer to a subscriber is increased, then  it is more likely that the subscriber will accept the offer. So, the probability of offer acceptance by a subscriber is modeled by the exponential distribution with mean or rate of the distribution as the acceptance rate. So, the acceptance probability is expressed as $\beta_i=1-e^{-\gamma x_i}$, where $x_i$ is the incentive offered to the subscriber. 

The revenue can be generated from a subscriber  if the customer does not churn out. Note that a  revenue $p_i$ is generated from the subscriber $i$  when it is sure that the subscriber does not churn out. In the churn scenario, the subscriber stays in the network with probability $1-\alpha_i$ and pays $p_i$. 
So, the expected revenue from $i^\text{th}$ subscriber is $(1-\alpha_i)p_i$.

Now consider the case when an offer is made to the subscriber who may accept the offer with probability $1-\beta_i$ or may reject the offer with probability $\beta_i$.
If he does not accept the offer, the revenue is $(1-\alpha_i)p_i$ with probability $1-\beta_i$.
If he accepts $x_i$ as an offer, then the revenue is $p_i-x_i$ with probability $\beta_i$
So, the expected revenue from the $i^\text{th}$ subscriber is  
\begin{equation}
f(x_i; \alpha_i, \gamma_i, p_i)= \beta_i(p_i-x_i) + (1-\beta_i)(1-\alpha_i)p_i
\end{equation}

The possible values of the offers may be non-negative integers or some predefined discrete values.
Suppose there are $k$ different types of offers and  the $j^\text{th}$ type has an offer value  $\delta_j$ and it can be awarded to  $n_j$ subscribers. 
So, possible discrete offer denominations are $\{ \delta_1,\delta_2,\ldots,\delta_k\}$.
Then the total value of type~$j$ offers is $w_j=\delta_jn_j$ and the total number of offers is $K=n_1+n_2+\ldots+n_k$, and hence, the total budget is $W=w_1+w_2+\ldots+w_k$.
Further, the unknown offer value $x_i$ for $i^\text{th}$ subscriber can be represented as  $x_i=\sum_{j=1}^k\delta_jx_{i,j}={\bf \delta}\cdot {\bf x}_i$, 
where vectors ${\bf \delta}=[\delta_1,\delta_2,\cdots,\delta_k]$ and  ${\bf x}_i=[x_{i,1},x_{i,2},\cdots,x_{i,k}]$ 
and each $x_{i,j}$ is a binary variable such that $x_{i,j}=1$ implies the $j^\text{th}$ offer is selected for $i^\text{th}$ subscriber. 
Note that the revenue function can be written as $f({\bf \delta}\cdot {\bf x}_i; \alpha_i, \gamma_i, p_i)=f(x_i; \alpha_i, \gamma_i, p_i)= \beta_i(p_i-x_i) + (1-\beta_i)(1-\alpha_i)p_i$. Then the offer optimization problem can be stated as  follow:
\begin{equation}
\begin{aligned}
\max \quad & F({\bf x}) =\sum_{i=1}^n f({\bf \delta}\cdot {\bf x}_i; \alpha_i, \gamma_i, p_i) \\
\textrm{S.t.} \quad & \\
  &\sum_{j=1}^kx_{i,j}\leq 1,~\forall i=1,2,\ldots,n \\
  &\sum_{i=1}^n x_{i,j}\leq n_j,~\forall j=1,2,\ldots,k\\
  &x_{i,j}\in \{0,1\},~\forall~ i ~\&~j\\
\end{aligned}
\end{equation}\label{eq:optprob}

 where the unknown variables are given by ${\bf x} =[{\bf x}_1,{\bf x}_2,\cdots,{\bf x}_n]^T$, ${\bf x}_i=[x_{i,1},x_{i,2},\cdots,x_{i,k}]$, and 
 the acceptance probability is computed as $\beta_i=1-e^{-\gamma_i\sum_{j=1}^k \delta_jx_{i,j}}$. 
However, this probability may be estimated by using different types of distribution  and the algorithm is equally applicable for optimizing the function $F({\bf x})$. 
A subscriber may receive at most one of the possible offers, which is enforced by the first constraint of the optimization problem. The second constraint ensures that the number of specific type of offers never exceed the number of available offers of the type. The third constraint implies that the decision variables are binary.

In our problem formulation, the restrictions on the budgets for each type of offer (i.e., $w_j$'s) and the total budget ($W$) are implicit and ensured by the given constraints. 
Hence, we have ignored  additional constraints such as $\sum_{i=1}^n \delta_jx_{i,j}\leq w_j, ~~\forall j=1,2,\cdots,k$ and $\sum_{j=1}^k\sum_{i=1}^n \delta_jx_{i,j}\leq W$ in the optimization.

Note that if the number of subscribers, $n$ is large (i.e., in order of million), the number of binary variables $nk$ is very large.
Naturally it is very difficult to to solve such a problem efficiently and accurately. 
In this study, we propose an efficient greedy algorithm for solving this problem.

\section{Proposed Algorithm}\label{se:algo}
The problem stated in Sec.~\ref{se:prob} is a very large scale optimization as the number of subscribes may be in the order of  million.
Solving this problem by a general optimization technique, in general, is difficult and inefficient.
Here we propose a greedy algorithm  which is  simple, elegant, and efficient for finding an optimal solution to the problem.

A greedy algorithm  finds an optimal solution based on some ``local optimal criteria" that leads to a global optimal solution. 
We would like to maximize the objective function by offering one subscriber at a time and the subscriber is selected by a greedy choice.
Let us take a closer look at the objective function $F({\bf x})=\sum_if(x_i; \alpha_i, \gamma_i, p_i)$, where each $f(x_i; \alpha_i, \gamma_i, p_i)$ is non-negative function, and it defined on a discrete set of values. 
We expect to maximize each of these functions in order to maximize $F({\bf x})$. 
Here we adopt the following greedy choice: \\

\noindent
{\sc Greedy Choice / Selection}:
{\em Choose a subscriber $i$ and an offer value $\delta_j$ from the available set of offer such that $f(\delta_j; \alpha_i, \gamma_i, p_i)$ is maximum among all available alternatives}. \\

\noindent
So, every time we choose the one that provides the maximum revenue and we continue till there is no more offer or all subscribers received the offer.

Suppose initially we have the set of subscribers, $S=\{1,2,\cdots,n\}$ to whom the offers will be made and let $A$, initialized to $A=\emptyset$, denote the set of subscribers who already have received an offer. 
We consider $k$ buckets, each containing a set of offers with same value.
Our approach would be to select a subscriber $i$ from the set of unassigned subscribers such that $f(x_i; \alpha_i, \gamma_i, p_i)$ is maximum for some offer, say $x_i=\delta_j$. 
This offer can be written as vector ${\bf x}_i=[x_{i,1}=0,x_{i,2}=0,\cdots, x_{i,j}=1, x_{i,j+1}=0,\cdots,x_{i,k}=0]$, and $x_i={\bf x}_i\cdot\delta$.
 As soon as we find such a subscriber $i$ and an offer $\delta_j$, we assign the offer  to the subscriber $i$, and remove it from $S$, remove the offer from $j$th bucket and we put $(i, j)$ in $A$. 
Next, from the remaining set of subscribers in $S$, we choose a subscriber  $l$  and an offer $\delta_{m}$ from the available set of offers such that the revenue function of the subscriber is maximum, i.e.,  $f(x; \alpha_l, \gamma_l, p_l)$ is maximum at $x=\delta_m$.
Then we remove the offer from the $m^\text{th}$ bucket and  the subscriber $l$ from $S$, and put $(l, m)$ into $A$. 
This process is continued until all subscribers are assigned with an offer or all the offer buckets are empty. 
If buckets are empty before $S$ is empty, then the remaining subscribers will not receive any offer.
For unified interpretation, we assume each unassigned subscriber receives a zero offer.

\subsection{Data Structures}
For an efficient implementation of the proposed algorithm, we need appropriate data structures. 
For a speedy execution of the greedy choice, we  exploit the priority queues of the revenue values $f(x; \alpha, \gamma, p)$ for each subscribers.
In order to make a greedy choice for a subscriber and an offer, we need to determine  $\max_j\{\max_i f(\delta_j; \alpha_i, \gamma_i, p_i)\}\}$.
For a given offer type~$j$, we construct a {\em max-priority queue} $Q_j$ of subscribers with the priority values as the revenue from the subscriber. So, the priority value for $i^\text{th}$ subscriber is $f(\delta_j; \alpha_i, \gamma_i, p_i)$.
Since, there are $k$ possible types of offers, we shall maintain $k$ priority queues, namely $Q_1,Q_2,\ldots,Q_k$.
Each of these priority queues is implemented by using binary heap, which is a complete binary tree stored in an array. 
The space complexity of a binary heap or priority queue is  $O(n)$ and it can be constructed in $O(n)$ time., where $n$ is the number of elements.
The front element or the max-priority element of a priority queue lies at the root of the underlying binary heap, and 
hence  it can be found in $O(1)$ time. 
Since deletion of an element from the queue requires  $O(\log n)$ time in an $n$-element queue,
 additional $O(\log n)$ time is required to maintain the queue after removal of the max-priority element. 

We also maintain an array $L$ of size $k$ that stores the references to the front elements of the priority queues $Q_1,Q_2,\ldots,Q_k$. 
At any point of time, we simply scan this array to find the elements (i.e., the subscriber) having the maximum priority value among all front elements of the queues. So, finding maximum of maximums can be done in $O(k)$ time.
Once this maximum is found, corresponding element is deleted from the queues and the array $L$ is also updated accordingly.
For details of binary heap and priority queue, and their properties please refer to Cormen~et.~al~\cite{cormen_09}.

We also use a lookup table for performing deletion of an element from $k$ priority queues efficiently.
The greedy choice determines a subscriber $i$ and an offer type $j$ such that subscriber $i$ lies at the front of $Q_j$.
So, while deleting corresponding queue element from $Q_j$ is easy and efficient as we know the position of the element in $Q_j$.
However, for the subscriber $i$, there are corresponding elements in the queues other than $Q_j$, but their positions are unknown, and hence, 
their deletions are not straightforward for achieving logarithmic time complexity.
In order to achieve efficient deletion of an element from $k$ priority queues, we use a lookup table $T$, which is a 2-dimensional array of dimension $n\times k$ storing references of each subscriber  in the priority queues. 
To be more precise, the $(i,j)^\text{th}$ element $T_{i,j}$ in $T$ stores the reference of $i^\text{th}$ subscriber in $j^\text{th}$ queue $Q_j$.
The lookup table is updated according to the changes performed in the priority queues.
Although this increases the computational overhead of the algorithm, but the time complexities of the operations on priority queues remain unchanged, and update operation on a queue can be performed in $O(\log n)$ time.
Implementation of $T$ requires $O(kn)$ memory and $O(kn)$ construction time.

\begin{figure}[!t]\begin{center}
\includegraphics[width=0.48\textwidth]{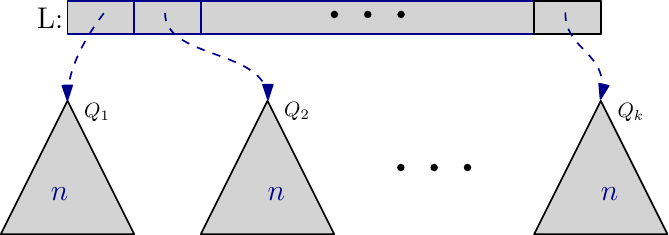}
\end{center}
\caption{Main data structures for implementing the proposed greedy algorithm. $Q_1,Q_2,\cdots,Q_k$ are max-priority queues of the subscribers. 
The queue $Q_j$ is a Max. Priority Queue and contains at most $n$ subscribers with $f(\delta_j;\cdot,\cdot,\cdot)$ as their priority values. 
The references to the roots of $Q_1, Q_2, \ldots, Q_k$ (i.e., max. priority elements) are stored in another list $L$ which helps in finding the maximum of maximum.}
\label{fig:queue}
\end{figure}

\vspace*{3mm}
\subsection{Algorithm Greedy Offer(AGO)}

The main steps of the proposed greedy algorithm is presented in Algorithm {\sc GreedOffer}, where  $S$ denotes the set of subscriber,  
$K=\sum_{j=1}^k n_j$  denotes  the total number of offers, and 
$A$ denotes the set of $2$-tuple $(i,j)$, which represent assignment of an offer of type $j$ to a subscribers $i$.
First initialization of $S$, $A$, and constructions of the max-priority queues $Q_1,Q_2,\cdots,Q_k$ are performed in Line~$1$~to~$2$.
Note that the front element of $Q_j$ has the maximum key value along all key values of the  elements stored in $Q_j$.
In the loop at Line~3, the algorithm selects  subscribers and assigns offer  iteratively by following the ``Greedy Choice/Selection" strategy earlier.
For example, if the greedy choice provides a subscriber $i$ and an offer type $j$, i.e., $i^\text{th}$ subscriber corresponds to the maximum of the key values stored in the $j^\text{th}$ priority queue $Q_j$, then  $f(\delta_j; \alpha_i, \gamma_i, p_i)=\max_{l,m}f(\delta_m; \alpha_l,\gamma_l,p_l)$ and it is the maximum over all subscribers and all possible offers. 
Then the algorithm inserts $(i,j)$ into $A$, and deletes $i^{th}$ subscriber from each of the queues  $Q_1,Q_2,\cdots,Q_k$ and reduce the value of $n_j$ by one, and reduce the total number of available offers $K$ by one.
Note that if $n_j$ is zero, no more offer of type-$j$ can be assigned to any of the remaining subscribers, and hence the queue $Q_j$ is no longer needed and hence, it is deleted. 
In the next iteration, the same steps are repeated for making a greedy choice, and continued till all offers are assigned to subscriber or no  subscriber is left. 
The steps of the  algorithm are described below.

\noindent
{\em  Algorithm {\sc GreedyOffer}($\alpha[1..n]$, $\gamma[1..n]$, $p[1..n]$)}
\begin{itemize}
\item [{ 1.}] Set $S=\{1,2,\ldots,n\}$, $A=\emptyset$ {\small\color{blue} // $O(n)$}
\item[{2.}] Construct  $Q_1,Q_2,\cdots,Q_k$ {\small\color{blue} //$O(kn)$}
\item[{3.}]{\bf while} ($K>0$\& $S\neq \emptyset$) {\bf do} {\small\color{blue} //repeats $n$ or $\sum n_j$ times}
\item[{4.}]\hspace*{4mm}$(i, j)\leftarrow$\,{\sc FindMaxOfMax}($Q_1,\ldots,Q_k$) {\small\color{blue}//$O(k)$}
\item[{5.}]\hspace*{4mm}$A\leftarrow A\cup\{(i,j)\}$  {\small\color{blue} //$O(1)$}
\item[{6.}]\hspace*{4mm}$S \leftarrow S\setminus \{i\}$  {\small\color{blue} //$O(1)$}
\item[{7.}]\hspace*{4mm}Delete $i$ from $Q_1,Q_2,\cdots,Q_k$ {\small\color{blue}//$O(k\log n)$}
\item[{8.}]\hspace*{4mm}$n_j \leftarrow n_j-1$, $K \leftarrow K-1$  {\small\color{blue} //$O(1)$}
\item[{9.}]\hspace*{4mm}{\bf If} ($n_j==0$) {\bf do}  {\small\color{blue} //$O(1)$}
\item[{10.}]\hspace*{9mm} Delete $Q_j$  {\small\color{blue} //$O(1)$}
\item[{11.}] {\bf return} $A$
\end{itemize}

Observe that the proposed algorithm is designed in such as way that a subscriber can receive at~most one offer from the available set, which is insured by the steps shown in line~$6$ to line~$10$. Also steps in line~$8$ to line~$10$ ensure that if a specific type of offers finishes,  none of the remaining subscribers receive such offer. So, the constraints of the optimization problem stated in Eq.~\ref{eq:optprob} are satisfied.

\subsection{Time and Space Complexity}\label{se:complex}
A priority queue of $n$ elements can be constructed in $O(n)$ time, and hence $k$ priority queues can be constructed $O(kn)$ time.
In Algorithm {\sc GreedyOffer},  the initialization of the data structures lines~1~to~2 can be executed in $O(kn)$ time.
The loop in line~5 is executed  $\max\{n, m\}$ times, where total number of offers is $m=\sum_{j=1}^k n_j$.
In general, $m<n$, and hence we can assume that this loop is executed $m$ times. 
Since, in line~9, an arbitrary subscriber can be deleted from a priority queue in  $O(\log n)$ time, the execution time of the entire loop consumes  $O(km\log n)$ time which is bounded from above by $O(kn\log n)$. 
Hence, the time complexity of the algorithm is $O(kn\log n)$.

Each priority queue contains at most $n$ elements, one for each subscriber, and each element can be stored using constant amount of space.
So the space complexity for storing a single queue is $O(n)$.
Since there are $k$ priority queues, one for each type of offer, we need $O(nk)$ memory for maintaining the queues.
So, the space complexity of the algorithm is $O(nk)$.

\subsection{Optimality of Solution}\label{se:prop}
We assume that  each subscriber is independent from others, i.e., the revenue from a subscriber depends only on the incentive offered to the subscriber, and it is independent of what incentives given to others.
 Under this assumption, the proposed algorithm provides an optimal solution to the optimization problem.
 The revenue function discussed in Section~\ref{se:prob} satisfies this condition. 
In order to prove our claim, we shall utilize the following observation about the greedy choice made by our algorithm.

\begin{obs}[Monotonicity]\label{obs:mono}
Suppose the greedy algorithm determines the offers $x_{i_1}, x_{i_2},\ldots,x_{i_n}$ to the subscribers in the order ${i_1}, {i_2},\ldots,{i_n}$ respectively. 
Then the corresponding revenues from the subscribers are in decreasing order, i.e.,\\
\begin{equation}
\begin{tabular}{ll}
 $f(x_{i_1}; \alpha_{i_1},\gamma_{i_1},p_{i_1})$ &$ \geq f(x_{i_2};  \alpha_{i_2},\gamma_{i_2},p_{i_2})$\\
&$\geq \ldots$ \\
& $\geq f(x_{i_n}; \alpha_{i_n},\gamma_{i_n},p_{i_n}).$
\end{tabular}
\end{equation}
\end{obs}

The proof of Observation~\ref{obs:mono} trivially follows directly from the fact that the greedy algorithm always chooses the subscriber who is giving maximum revenue under the available offers at that moment.

\begin{theorem}[Optimality]
The solution provided by the greedy algorithm is optimal.
\end{theorem}
{\sc Proof:} Let ${\bf X}=[x_1,x_2,\ldots,x_n]$ be a solution given by the algorithm. Let ${\bf Y}=[y_1,y_2,\ldots,y_n]$ be an arbitrary solution. The revenue corresponding to the offer vectors ${\bf X}$ and ${\bf Y}$, respectively  are given by:
\begin{equation}
\begin{tabular}{@{}c@{}l@{}}
$F({\bf X})=$&$\sum_{i=1}^n f(x_i; \alpha_i, \gamma_i, p_i) = \sum_{i=1}^nf_i(x_i)$ and \\
$F({\bf Y})=$&$\sum_{i=1}^n f(y_i; \alpha_i, \gamma_i, p_i)=\sum_{i=1}^nf_i(y_i)$,
\end{tabular}
\end{equation}
where, for simplicity, we denote$ f_i(x) := f(x; \alpha_i, \gamma_i, p_i)$.
In order to prove that the greedy algorithm provides an optimal solution, it sufficient to show that $F({\bf X})\geq F({\bf Y})$.

Without loss of generality, we assume that the subscribers are re-arranged in such a way that the algorithm first finds the offer $x_1$ for subscriber $1$, then $x_2$ for subscriber  $2$, and so on, and finally $x_n$ for $n$. 
Note that some of these offers may be zero.
Let $i$ be the smallest index with $x_i\neq y_i$, i.e, $x_1=y_1,x_2=y_2,\ldots,x_{i-1}=y_{i-1}, x_i\neq y_i$. 

We have 
\begin{equation}\label{eq:revdiff1}
\begin{tabular}{@{}@{}l@{}@{}l@{}}
$F({\bf X})-F({\bf Y})$&\\
$~~=\sum_{i=1}^n f_i(x_i) - \sum_{i=1}^n f_i(y_i)$&\\
$~~=\sum_{i=1}^n\left( f_i(x_i)-f_i(y_i)\right)$, { \text{since}  $x_j=y_j~ \forall j< i$}&\\
$~~=(f_i(x_i)-f_i(y_i)) + (f_{i+1}(x_{i+1})-f_{i+1}(y_{i+1}))$& \\
$~~~~~ + \ldots + (f_n(x_n)-f_n(y_n)).$&
\end{tabular} 
\end{equation}
According to the greedy algorithm, $x_i$ is the best available offer made to the $i^{th}$ subscriber such that the additional revenue coming from the $i^{th}$ subscriber is maximum. i.e., 
\begin{equation}\label{eq:revdiff2}
f_i(x_i) = \max_{z_j\in O_i , k\in S_i} f_k(z_j) \geq f_i(y_i)
\end{equation}
where $O_i$ is the set of all remaining offers and $S_i$ is the set of remaining subscribers to be offered at the time of the $i^\text{th}$ greedy choice made by the algorithm.
Then by Eq.~\ref{eq:revdiff1} and Eq.~\ref{eq:revdiff2},
it follows that $f_{i+1}(x_{i+1})\geq f_{i+1}(y_{i+1})$,  $f_{i+2}(x_{i+2})\geq f_{i+2}(y_{i+2})$,  $\ldots$,  $f_n(x_n)\geq f_n(y_n)$. Hence, $F({\bf X})-F({\bf Y})\geq 0$, i.e., $F({\bf X})\geq F({\bf Y})$.  This completes the proof.

Finally, from the discussions in Sec.~\ref{se:complex} and Sec.~\ref{se:prop}, the novelty of the proposed algorithm can be summarized in the following theorem.
\begin{theorem}
Algorithm {\sc GreedyOffer}  determines an optimal solution 
the offer optimization problem (given in Eq.~\ref{eq:optprob}) in  $O(kn\log n)$ time, and it consumes $O(kn)$ memory for solving the problem.
\end{theorem}

\begin{table}[!thb]\center
\caption{Example of hypothetical non-monetary offers.}\label{tab:nmoffer}
\begin{tabular}{@{\,}c@{\,}|@{\,}c@{\,}|@{\,}c@{\,}|@{\,}c@{\,}}
\hline
{\bf Offer Name} & {\bf Value $\delta_j$\$}   & {\bf Count $n_k$} & {\bf Sum $w_j=\delta_jn_k$ \$}\\\hline
5GB Data    &  $5$            &  $90$      &  $450$\\\hline
Hotspot 2GB&  $2$            &  $50$      &  $100$\\\hline
100 SMS     &  $4$            &  $80$      &  $320$\\\hline
5 Movies     &  $8$            &  $70$      &  $560$\\\hline
$\vdots$     & $\vdots$      &  $\vdots$ &  $\vdots$     \\\hline
      &    &  $K=\sum_j n_j$& $W=\sum w_j$\\\hline
\end{tabular}
\end{table}

\subsection{OOP with Heterogeneous Offers}
The problem stated in Sec.~\ref{se:prob} is designed with  offers in terms of monetary incentive.
The same is also applicable to a more generalized scenario where the offers are made in terms of services and non-monetary offers such as additional data, hot-spot tethering, additional SMS, value added services or specific download limit, etc. 
These additional services or non-monetary offers can be represented in terms of equivalent monetary advantages. 
An example of such offers are shown in Table~\ref{tab:nmoffer}, in which  offers are presented as different types with equivalent values, and hence, we  solve it by using the same formulation as stated in Sec.~\ref{se:prob}. 
In order to incorporate the subscriber specific profile or usage history of a subscriber, a relative weight to each type of offers can also be set for incorporating in the objective function.


\begin{table*}[ht]
\caption{Performance comparison proposed algorithm with Genetic algorithm available  in the python library {\em pymoo}~(\cite{deb2020,pymoolib}) and a non-linear solver IPOPT available in the  python library {\em pyomo}~(\cite{hart2011pyomo,bynum2021pyomo}). All algorithms are executed on simulated data to find the maximum value of the objective function and the corresponding offer distribution.}
\begin{center}
\begin{tabular}{ccc|cccc|cc}
   \hline
    ${\bf\large  n}$& \multicolumn{2}{c|}{\bf Proposed~Algorithm} & \multicolumn{4}{c|}{\bf  Genetic Algorithm} & \multicolumn{2}{c}{\bf  Pyomo IPOPT}\\
    & {\bf Opt.\,Val.} & {\bf Time\,(sec.)} & {Pop.\,size} & {Iter.} & {Opt. Val.} & { Time\,(sec.)} & {Opt.\,Val.} &{ Time\,(sec.)}\\\hline
$~~100$&${\bf ~~3676.81}$&${\bf 0.0469}$ & $200$ & $500$ & $~~3565.14$  & $1009.5625$ & $~~3608.85$ & $602.1000$\\ 
$~~200$&${\bf ~~7191.99}$&${\bf 0.0625}$ & $200$ & $500$ & $~~6729.98 $&$1589.0938$ &$~~7040.70$&$602.3029$\\\hline 
$~~300$&${\bf 11825.49}$&${\bf 0.1094}  $& $200$  & $500$  & $11216.57$ &$2198.7813$  & $11753.27$& $604.4841$\\
$~~400$&${\bf 15285.47}$&${\bf 0.1406}  $& $200$  & $500$  & $14278.32$  &$3276.4375$  &$14976.10$ &$604.6450$\\\hline
$~~500$&${\bf 18556.30}$&${\bf 0.1875}  $& $300$  & $600$  & $17682.03$ &$9310.2188$ &$18309.22$ & $606.1524$\\ 
$~~600$&${\bf 23353.26}$&${\bf 0.1889}  $& $300$  & $600$  & $20747.12$ & $11443.1406$ & $22789.53$&$607.9732$\\\hline
$~~700$&${\bf 25780.47}$&${\bf 0.2031}  $&$500$   & $700$  & $23804.25$  &$20463.4062$ &$24824.68$&$606.5911$\\ 
$~~800$&${\bf 29108.80}$&${\bf 0.2344}  $&$600$   & $800$  & $26931.79$  &$30028.0000$ &$27512.51$ & $606.8904$\\\hline 
$~~900$&${\bf 33884.70}$&${\bf 0.3594}  $&$600$   & $900$  & $31777.81$  &$38664.6406$ &$33742.06$ & $608.2689$ \\
   $1000$&${\bf 37909.31}$ & ${\bf 0.5000}$&$600$   & $900$  &$34911.58$  &$42032.8125$  &$37169.06$ & $616.4576$\\\hline 
\label{tab:exetime}
\end{tabular}
\end{center}
\end{table*}

\section{Experimental Results}\label{se:exp}
All experimental results are obtained by using as noted book computer with  Intel(R) Core(TM) i5~\@2.60GHz processor, 16GB RAM, and Windows~10 Enterprise OS.
In order to show the novelty of the proposed algorithm, we present the experimental results in two parts. 
\begin{itemize}
\item Comparison performance with two standard algorithms
\item Efficiency and capability of proposed algorithm for handling large problems.
\end{itemize}

\subsection{Implementation}
The proposed algorithm is implemented in pure Python and no special library is used for the implementation.
The implementation of max-heap and  max-priority queue is done by following the standard algorithm  in \cite{cormen_09}.
We  have inserted additional steps required for construction of priority queues, deletion key and update operation in the priority queues in order  to maintain the consistency of the positions of a subscriber in the queues and in the lookup table adopted in the proposed algorithm. Although the maintenance of lookup table brings a small computational overhead, it makes the deletion and update operation very fast (i.e., logarithmic time).

\subsection{Comparison of Performance}
The efficiency (i.e., execution time) of the proposed algorithm is compared with a couple of standard optimization algorithms, namely, i)~{\em Genetic Algorithm} with constraints  that is available in the python library {\em pymoo}~(\cite{deb2020}), and ii)~constraints non-linear solver IPOPT available in the python library {\em pyomo}~(\cite{hart2011pyomo,bynum2021pyomo}).
In general, for problems with large number of unknowns, these techniques either take long time to find a good solution, or fail to converge to a feasible solution.
So, the comparison is shown with moderate size of the problem where number of subscribers varies from $100$ to $1000$ with the second parameter $k$ remains fixed to $5$. 
So, the number of decision variables for corresponding optimization problems varies for $nk=500$ to $nk=5000$.

In general, the genetic algorithm demands  large population size and large number of iterations, otherwise it cannot find a feasible solution for large number of unknowns.
So, for conducting the experiment, the population size and the number of iterations are increased as the problem size increases. The proposed algorithm completes execution in less than a second from finding an optimal solution solution, which is negligible compared to the time consumed by genetic algorithm.

Note that the pyomo solver also consumes significantly longer time than the proposed algorithm for solving the same problem and yet the optimal value of the objective function provided by it is smaller than the one provided by the proposed algorithm.

The execution time and the maximum value of the objective functions for problems with varying size are tabulated in Table~\ref{tab:exetime}.
This result clearly shows that the proposed algorithm always provides largest value of the objective function, while taking smallest amount of execution time. 
This is expected as these algorithms, in general, provides a sub-optimal solution to a problem, while our algorithm always provides an optimal solution.

\subsection{Efficiency with Large Problems}
\begin{figure}[!ht]\begin{center}
\includegraphics[width=0.46\textwidth]{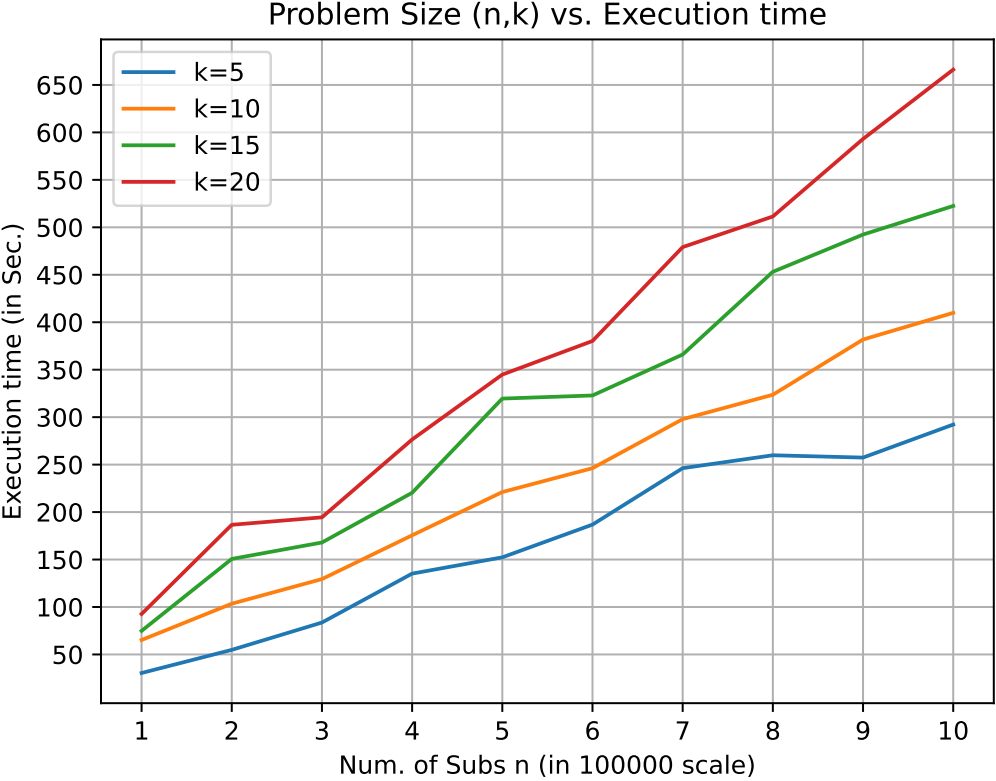}
\end{center}
\caption{Execution times (in sec.) consumed by the proposed algorithm for inputs with varying number of subscribers ($n$) from $100$ thousand to one million  are shown here, where each curve shows the results for $k=5,10,15,20$,  which the number of types of offers.}
\label{fig:exegraph2}
\end{figure}

In order to show the capability of handling large scale optimization problem, we have also furnished experimental results with very large number of subscribers which ranges from $100$ thousands to one million, and offer types varies from $5$ to $20$. The execution time is tabulated in Table~\ref{tab:exetime2} and 
the growth of the execution time is also visualized in Fig.~\ref{fig:exegraph2}, which shows that the execution time grows slowly, and it is  similar to a linear or log-linear function. 
So, the empirical results does not deviate from the theoretical bound of time complexity of our algorithm.

\begin{table}[!ht]
\caption{Execution time (in sec.) of proposed algorithm for different types of offers $k=5, 10, 15, 20$ and increasing number of subscribers starting from $n=100000$ to $n=1000000$.}
\begin{center}
\begin{tabular}{ccccc}
   \hline
$n$ & $k=5$ & $k=10$ & $k=15$ & $k=20$\\\hline
 $100000$&   $30.453$ & $65.217$& $74.844$&$   92.6406$\\
 $200000$&  $ 54.828$ & $103.469$& $ 150.625$&$ 186.594$\\\hline
 $300000$&   $83.625$ & $129.406$& $ 167.953$&$  194.453$\\
 $400000$& $135.172$ & $175.531$& $ 220.344$&$  276.460$\\\hline
 $500000$& $152.328$ & $221.109$& $ 319.578$&$  344.844$\\
 $600000$& $186.797$ & $246.234$& $ 322.844$&$  380.250$\\\hline
 $700000$& $246.250$ & $297.891$& $ 366.000$&$  479.156$\\
 $800000$& $259.859$ & $323.609$& $ 453.141$&$  511.422$\\\hline
 $900000$& $257.406$ & $381.781$& $492.375$&$   593.094$\\
$1000000$& $292.203$& $409.906$& $ 522.500$&$   666.156$\\\hline
\label{tab:exetime2}
\end{tabular}
\end{center}
\end{table}

\section{Conclusion}\label{se:con}
Here we conclude with few remarks on applications, efficiency, improvement, and shortcomings.
\begin{itemize}
\item In this paper, we proposed a greedy algorithm for solving a special type of combinatorial optimization problem, namely offer optimization problem that comes from telecom domain.
However, the algorithm may be used for solving similar problem coming from different domains, which requires allocation and utilization of resources to optimize predefined objectives.
\item The time complexity of the algorithm can be improved by using another priority queue $Q$ for implementing the greedy choice of the proposed algorithm, which compute the maximum of maximums for selecting a subscriber and an offer that make maximum revenue. Instead of the list $L$, if we use another queue $Q$ to storing the references of the roots of $k$ priority queues, then the greedy choice can be made in $O(\log k)$ time instead of $O(k)$ time.

\item The efficiency of the algorithm is achieved with a cost of memory. 
The space complexity of the algorithm is $O(kn)$ which not a big issue if number of offer types $k$ is of moderate size. The typical  values of $k$ in telecom domain may be  $5$ to $20$. 
For problem coming from some domain such as online retailer, the number of different types of offer may be large, and then the memory requirement will be increased.
The algorithm may be extended for the environment of distributed computing to handle such issues as well as increasing the efficiency.
\end{itemize}

\bibliography{uai2023ref}

\begin{thebibliography}{11}
\providecommand{\natexlab}[1]{#1}
\providecommand{\url}[1]{\texttt{#1}}
\expandafter\ifx\csname urlstyle\endcsname\relax
  \providecommand{\doi}[1]{doi: #1}\else
  \providecommand{\doi}{doi: \begingroup \urlstyle{rm}\Url}\fi

\bibitem[pym(2020)]{pymoolib}
Multi-objective optimization in python.
\newblock \url{https://pymoo.org/}, 2020.
\newblock Accessed: 12-02-2023.

\bibitem[{Blank} and {Deb}(2020)]{deb2020}
J.~{Blank} and K.~{Deb}.
\newblock pymoo: Multi-objective optimization in python.
\newblock \emph{IEEE Access}, 8:\penalty0 89497--89509, 2020.

\bibitem[Bynum et~al.(2021)Bynum, Hackebeil, Hart, Laird, Nicholson, Siirola,
  Watson, and Woodruff]{bynum2021pyomo}
Michael~L. Bynum, Gabriel~A. Hackebeil, William~E. Hart, Carl~D. Laird,
  Bethany~L. Nicholson, John~D. Siirola, Jean-Paul Watson, and David~L.
  Woodruff.
\newblock \emph{Pyomo--optimization modeling in python}, volume~67.
\newblock Springer Science \& Business Media, third edition, 2021.

\bibitem[Cohen(2004)]{cohen2004exploiting}
M-D Cohen.
\newblock Exploiting response models—optimizing cross-sell and up-sell
  opportunities in banking.
\newblock \emph{Information Systems}, 29\penalty0 (4):\penalty0 327--341, 2004.

\bibitem[Cormen et~al.(2009)Cormen, Leiserson, Rivest, and Stein]{cormen_09}
Thomas~H. Cormen, Charles~E. Leiserson, Ronald~L. Rivest, and Clifford Stein.
\newblock \emph{Introduction to Algorithms}.
\newblock The MIT Press, 3rd edition, 2009.
\newblock ISBN 0262032937.

\bibitem[Hart et~al.(2011)Hart, Watson, and Woodruff]{hart2011pyomo}
William~E Hart, Jean-Paul Watson, and David~L Woodruff.
\newblock Pyomo: modeling and solving mathematical programs in python.
\newblock \emph{Mathematical Programming Computation}, 3\penalty0 (3):\penalty0
  219--260, 2011.

\bibitem[Johnson et~al.(2013)Johnson, Tellis, and Ip]{johnson2013whom}
Joseph Johnson, Gerard~J Tellis, and Edward~H Ip.
\newblock To whom, when, and how much to discount? a constrained optimization
  of customized temporal discounts.
\newblock \emph{Journal of Retailing}, 89\penalty0 (4):\penalty0 361--373,
  2013.

\bibitem[Nobibon et~al.(2011)Nobibon, Leus, and
  Spieksma]{nobibon2011optimization}
Fabrice~Talla Nobibon, Roel Leus, and Frits~CR Spieksma.
\newblock Optimization models for targeted offers in direct marketing: Exact
  and heuristic algorithms.
\newblock \emph{European Journal of Operational Research}, 210\penalty0
  (3):\penalty0 670--683, 2011.

\bibitem[Pham et~al.(2021)Pham, Chu, Pham, Dao, Pham, Nguyen,
  et~al.]{pham2021recommendation}
Cong~Dan Pham, Tuan~Anh Chu, Huy~Hung Pham, Manh~Linh Dao, Thanh~Son Pham,
  Duc~Hai Nguyen, et~al.
\newblock A recommendation system for offers in telecommunications.
\newblock In \emph{2020 IEEE Eighth International Conference on Communications
  and Electronics (ICCE)}, pages 302--306. IEEE, 2021.

\bibitem[Resende and Pardalos(2008)]{resende2008handbook}
Mauricio~GC Resende and Panos~M Pardalos.
\newblock \emph{Handbook of optimization in telecommunications}.
\newblock Springer Science \& Business Media, 2008.

\bibitem[Verma(2020)]{verma2020offer}
Ankur Verma.
\newblock Offer personalization using temporal convolution network and
  optimization.
\newblock \emph{arXiv preprint arXiv:2010.08130}, 2020.

\end{thebibliography}
\end{document}